# A Universal Knowledge Model and Cognitive Architecture for Prototyping AGI

Artem Sukhobokov[1[0000-0002-1370-6905]], Evgeny Belousov[2[0009-0002-5335-9167]], Danila Gromozdov[2[0009-0007-2342-6046]], Anna Zenger[3[0009-0007-9909-755X]], and Ilya Popov[2*[0009-0007-2226-9640]]

[1] SAP America, Inc. 3999 West Chester Pike, Newtown Square, PA 19073, USA
[2] Bauman Moscow State Technical University, 2-nd Baumanskaya 5, Moscow 105005, Russian Federation
[3] OZON Marketplace Kazakhstan LLP, Al-Farabi avenue 77/7, Bostandyk district, Almaty 050040, Kazakhstan
link.s.twink@gmail.com

**Abstract.** The article identified 42 cognitive architectures for creating general artificial intelligence (AGI) and proposed a set of interrelated functional blocks that an agent approaching AGI in its capabilities should possess. Since the required set of blocks is not found in any of the existing architectures, the article proposes a new cognitive architecture for intelligent systems approaching AGI in their capabilities. As one of the key solutions within the framework of the architecture, a universal method of knowledge representation is proposed, which allows combining various non-formalized, partially and fully formalized methods of knowledge representation in a single knowledge base, such as texts in natural languages, images, audio and video recordings, graphs, algorithms, databases, neural networks, knowledge graphs, ontologies, frames, essence-property-relation models, production systems, predicate calculus models, conceptual models, and others. To combine and structure various fragments of knowledge, archigraph models are used, constructed as a development of annotated metagraphs. As components, the cognitive architecture being developed includes machine consciousness, machine subconsciousness, blocks of interaction with the external environment, a goal management block, an emotional control system, a block of social interaction, a block of reflection, an ethics block and a worldview block, a learning block, a monitoring block, blocks of statement and solving problems, self-organization and meta learning block.

## 1 Introduction

In recent years, there has been a rapid and multidirectional development of intelligent information systems developed by people [1]. In [2], it was assumed that a relatively distant extrapolation of the functionality of these developments would be the creation of a general artificial intelligence comparable to human intelligence (in current terminology – AGI). The main capabilities of an intelligent information system, such as perceptual abilities, attention mechanisms, choice of actions, learning, memory, reasoning and their practical application are determined by the cognitive architecture used in the system [3]. If we consider the basic principles of functioning, then cognitive architectures are the opposite of expert systems. Expert systems provide solutions to intellectual tasks in a narrowly defined context, in the segment of activity for which they have knowledge, in contrast, cognitive architectures aim to providing a wide coverage, solving a diverse set of tasks in different fields. More importantly, cognitive architectures provide intelligent behavior at the system level, rather than at the level of methods of individual components designed to solve specialized tasks [3].

In 2016-18, after a relatively long break, two reviews of cognitive architectures were published [4, 5] (the third version of the publication [4], like the publication [5], appeared in 1918). After that, no systematic reviews were published until preparation of this article. Only brief reviews were published as part of the justification for the need to develop new architectures. In these two reviews, which cover about 140 architectures, we identified cognitive architectures designed to create AGI, and supplemented the list with the results of a bibliographic search.

Unlike the reviews in [6,7], the authors of which considered cognitive architectures for creating AGI, including architectures that were not intended for this purpose in the review, we consider it important to analyze precisely cognitive architectures that are declared as architectures intended for creating AGI, since additional requirements are imposed on them [8]. Other cognitive architectures could be developed to test some technical solutions or to solve more utilitarian tasks, for example, for image processing or controlling a transport robot transporting workpieces in the workshop.

In the list of cognitive architectures, if the architecture does not have a name, it will be presented simply with a link to the publication. The final list of cognitive architectures for analysis includes 42 architectures, presented in Table 1. For each architecture, this table contains citations indicating that the architecture is intended to enable AGI. For some architectures, publications do not explicitly mention AGI as a target. In these cases, the decision to include in the list is determined by the goals of the architecture to model human-like behavior or by focusing on the complex of functions inherent with humans.

**Table 1.** Cognitive architectures aimed to create AGI, and author citations confirming this.

| # | Name | Link(s) | Citations about purpose of cognitive architecture |
|---|------|---------|---------------------------------------------------|
| 1 | Soar | [9] | Soar … have used … to build complex integrated AI agents, and to create detailed models of human behavior. … We have found that combining what is known in psychology, in neuroscience, and in AI is an effective approach to building a comprehensive cognitive architecture. … Our bet is that achieving human-level intelligence is a long path of incremental experiments, discoveries, tests, reformulations and refinements. |
| 2 | ACT-R | [10] | This paper explores requirements on cognitive architectures for artificial general intelligence. The goal of the analysis is to determine the requirements for cognitive architectures that support the full-range of human-level intelligent behavior. |

| 3 | NARS | [11] | … system aimed at the realization of Artificial General Intelligence (AGI). |
|---|---|---|---|
| 4 | LIDA | [12] | … we … argue that Learning Intelligent Distribution Agent (LIDA) … may be suitable as an underlying cognitive architecture on which others might build an AGI. |
| 5 | Haikonen cognitive architecture | [13] | The author visions autonomous robots that perceive and understand the world and act in it in a natural way, without programs and numerical representation of information. This approach considers the cognitive machine as a system that is seamlessly interactive, both internally and externally, in respect to its environment and experience. This approach should result in robots that know and understand what they are doing and why, robots that can plan and imagine their actions and the possible outcome of these. Robots that exhibit properties like these are said to possess machine consciousness, which may or may not have common deeper properties with animal and human consciousness. |
| 6 | SiMA (previously ARS) | [14] | … most of humans' behaviour is covered by every-day capabilities. … our experience with the cognitive architecture SiMA showed that – especially when the foundations of the human mind are at stake – every-day behaviour is more suitable to analyse the basic functions of the human mind. |
| 7 | Sigma | [15] | We … expect that a system capable of artificial general intelligence (AGI) would provide natural support for Theory of Mind. We are interested here in how Theory of Mind capabilities may be realized within Sigma ($\Sigma$), a nascent cognitive system—an integrated computational model of intelligent behavior— that is grounded in a cognitive architecture, a model of the fixed structure underlying a cognitive system. |
| 8 | | [16] | The platform used as an specific domain for the initial experiments is the iCub humanoid robot simulator, but the architecture is built so it can be applied to different platforms and applications. This platform was chosen because it provides a "Human-Like" architectural level … |
| 9 | CogPrime | [17,18] | Part 1 of the book … sketches the broad outlines of a novel, integrative architecture for Artificial General Intelligence (AGI) called CogPrime … Part 2 of the book concludes with a chapter summarizing the argument that CogPrime can lead to human-level (and eventually perhaps greater) AGI … |
| 10 | Ikon Flux | [19,20] | Ikon Flux is a fully implemented prototypical architecture for self-programming systems - a prototype being an abstract type to be instantiated in a concrete domain. … A system continuously modeling its own operation has to do so at multiple levels of abstraction, from the program rewriting up to the level of global processes (e.g. the utility function), thus turning eventually into a fully self-modeling system. ,,, We believe peewee granularity is a promising way to simplify operational semantics and reach a computational homogeneity that can enable automated architectural growth – which in itself is a necessary step towards scaling of cognitive skills exhibited by current state-of-the-art architectures. Only this way will we move more quickly towards artificial general intelligence. |
| 11 | eBICA | [21,22] | This work continues the effort to design and test the cognitive architecture eBICA: a general model of emotionally biased behavior control and decision making, with the focus on social emotional relationships. … We also presented the study of an implemented Virtual Actor based on the eBICA model. … The overall conclusion is that the implemented Virtual Actor performs at a human level. |
| 12 | D-LANCA | [7] | This work suggests a novel approach to autonomous systems development linking autonomous technology to an integrated cognitive architecture with the aim of supporting a common artificial general intelligence (AGI) development. |
| 13 | ICOM | [23] | This paper articulates the methodology and reasoning for how biasing in the Independent Core Observer Model (ICOM) Cognitive Architecture for Artificial General Intelligence (AGI) is done. |

| 14 |  | [24] | We consider a task-oriented approach to AGI, when any cognitive problem, perhaps superior to human ability, has sense given a criterion of its solution. |
|---|---|---|---|
| 15 | LIS | [25] | … the LIS Framework … provides a way to approach AGI learning in a flexible and easy-to-use manner by combining multiple, interchangeable components. The framework allows AGI workers including beginners to combine pre-installed LIS Framework components and begin AGI development with ease. |
| 16 |  | [26] | This paper is concerned with artificial general intelligence (AGI). Our ultimate goal is to create a computational model that may operate in any environment and develop intelligence adapted to that environment in a fully automatic fashion. |
| 17 | MBCA | [27] | The biologically inspired Meaningful-Based Cognitive Architecture (MBCA) integrates the sensory processing abilities found in neural networks with many of the symbolic logical abilities found in human cognition. … MBCA can functionally produce a variety of behaviors which can help to better hypothesize and understand mammalian cortical function, and provide insight into possible mechanisms which link such mesoscopic functioning to the causal and symbolic behavior seen in humans. |
| 18 |  | [28] | We introduce an AGI, in the form of cognitive architecture, which is based on Global Workspace Theory (GWT). |
| 19 | PySigma | [29] | The Sigma cognitive architecture is the beginning of an integrated computational model of intelligent behavior aimed at the grand goal of artificial general intelligence (AGI). However, whereas it has been proven to be capable of modeling a wide range of intelligent behaviors, the existing implementation of Sigma has suffered from several significant limitations. ... In this article, we propose solutions for this limitation ... The resulting design changes converge on a more capable version of the architecture called PySigma. |
| 20 |  | [30] | In general, neuromorphic computing and cognitive modeling are two promising approaches to achieve AGI. … Our motivation is to provide a generic methodology bridging the computation theory with the underlying implementation at the algorithm level. |
| 21 | GLAIR | [31] | GLAIR (Grounded Layered Architecture with Integrated Reasoning) is a multilayered cognitive architecture for embodied agents operating in real, virtual, or simulated environments containing other agents. … The motivation for the development of GLAIR has been "Computational Philosophy", the computational understanding and implementation of human-level intelligent behavior without necessarily being bound by the actual implementation of the human mind. |
| 22 |  | [32,33] | I'll be using the term my cognitive system, cognitive architecture, artificial general intelligence, and AGI interchangeably. |
| 23 | H-CogAff | [34] | H-CogAff, a special case of CogAff, is postulated as a minimal architecture specification for a human-like system. |
| 24 |  | [35] | Our approach is to use a functional core to simulate the development of cognitive functions of autonomous agents. … The most important goal in the field of AGI is the development of control systems for cognitive agents, which, in terms of their intellectual performance, are not inferior, and perhaps even surpass humans. Developmental psychology studies show that the most significant changes in Innenwelt occur at the sensorimotor stage, which is the first in the postnatal ontogenesis. … . According to Piaget, this period of development is one of the most important in the creation of human mental abilities. The proposed architecture makes it possible to simulate the process of evolution of cognitive abilities, including the stage of sensorimotor development of autonomous agents. |
| 25 | EM-ONE | [36] | The design of EM-ONE draws heavily on Minsky's Emotion Machine architecture hence the name EM-ONE … I have also drawn ideas from Sloman's |
| | | | H-CogAff architecture, which resembles Minsky's architecture in many respects ... Both Minsky and Sloman developed their architectures to provide rich frameworks with which to explain the diversity of complex and subtle aspects of human cognition, especially our capacity for common sense and our variety of emotions. ... My goal with EM-ONE is primarily to support more intricate forms of reflective commonsense thinking, although in the long run I hope it will help to explain a broader array of types of thinking including such feelings as love, confusion, anger, and hope. |
| 26 | | [37] | The article describes the author's proposal on cognitive architecture for the development of a general-level artificial intelligent agent («strong» artificial intelligence). |
| 27 | CAMAL | [38] | With the distributed model specific agents can change... The distributed agent requests those elements of itself (i.e. its component agents) that are associated with the current learning task to modify themselves. ... Currently we are looking at the nature of communication between agents with shared motivations and are using distributed blackboards. ... We have designed and implemented agents that display motivational qualities and address important questions about the nature of emotion and autonomy. ... We have demonstrated emotive qualities in our research agents. ... compromises can lead to the design of agent systems with inherent conflicts. ... By designing agents with the qualities described in this chapter an agent is given the means to represent and reason about these conflicts when they do arise. This research continues to raises questions about what agent is, what a mind is, and what are emotion and motivation. |
| 28 | SMCA | [39] | Artificial Intelligence originated with the desire to develop artificial minds capable of performing or behaving like an animal or person. ... Cognitive architectures are designed to be capable of performing certain behaviours and functions based on our understanding of human and non human minds. ... developing SMCA (Society of Mind Cognitive Architecture) can be viewed from the perspective of Minsky, which leads to the development of many different types of simple agents, with different behaviours. Metacognition is useful for framing the constraints for this swarm intelligence. Swarm intelligence requires the inclusion of a mathematical theory of how the group of agents work together to achieve a common goal. Swarm intelligence uses different mathematical algorithms so as to cover all processing and functioning associated with the adopted architecture or mind model. |
| 29 | MicroPSI | [40] | This book is completely dedicated to understanding the functional workings of intelligence and the mechanisms that underlie human behavior by creating a new cognitive architecture. |
| 30 | LISA | [41] | Human mental representations are both flexible and structured – properties that, together, present challenging design requirements for a model of human thinking. The Learning and Inference with Schemas and Analogies (LISA) model of analogical reasoning aims to achieve these properties within a neural network. |
| 31 | CIT | [42] | The paper proposes a novel cognitive architecture that combines cognitive computing and cognitive agent technologies for performing human-like functionality. The system architecture is known as CIT (Cognitive Information Technology). |
| 32 | Companion | [43] | The Companion cognitive architecture is aimed at reaching human-level AI by creating software social organisms – systems that interact with people using natural modalities, working and learning over extended periods of time as collaborators rather than tools. |
| 33 | | [44] | The paper proposes a novel cognitive architecture for computational creativity based on the Psi model and on the mechanisms inspired by dual process theories of reasoning and rationality. |

| 34 | Polyscheme | [45] | This thesis describes a new framework for understanding and creating human-level intelligence by integrating multiple representation and inference schemes. |
| 35 | | [46] | We present a three level Cognitive architecture for the simulation of human behaviour based on Stanovich's tripartite framework. |
| 36 | Clarion | [47] | The goal of this work is to develop a unified framework for understanding the human mind, and within the unified framework to develop process-based, mechanistic explanations of a substantial variety of psychological phenomena. |
| 37 | Oscar | [48] | The basic observation that motivates the OSCAR architecture is that agents of human-level intelligence operating in an environment of real-world complexity (henceforth, GIAs — "generally intelligent agents") must be able to form beliefs and make decisions against a background of pervasive ignorance. |
| 38 | OntoAgent | [49] | This paper presents an overview of a cognitive architecture, OntoAgent, that supports the creation and deployment of intelligent agents capable of simulating human-like abilities. |
| 39 | INKA | [50] | Artificial intelligence research is now flourishing which aims at achieving general, human-level intelligence. Accordingly, cognitive architectures are increasingly employed as blueprints for building intelligent agents to be endowed with various perceptive and cognitive abilities. This paper presents a novel integrated neuro-cognitive architecture (INCA) which emulate the putative functional aspects of various salient brain sub-systems via a learning memory modeling approach. |
| 40 | ISAAC | [51] | A foundational component of an ISAAC processing framework is the concept of "mixture of experts" architecture and methodology, similar to a human brain. |
| 41 | | [52] | This article provides an analytical framework for how to simulate human-like thought processes within a computer. ... Iterative updating is conceptualized here as an information processing strategy, a model of working memory, a theory of consciousness, and an algorithm for designing and programming artificial general intelligence. |
| 42 | Aigo | [53] | Here we outline an architecture and development plan, together with some preliminary results, that offers a much more direct path to full Human-Level AI (HLAI) / AGI. |

When analyzing the architectures, the following functional components were identified:

- Consciousness for agent management in real time;
- Subconscious mind for performing routine operations;
- Goal management;
- Emotional management;
- Formation and application of ethical assessments;
- Monitoring;
- Training;
- Social interaction;
- Reflection;
- Setting tasks;
- Problem solving;
- Self-development and meta-learning;
- Formation and use of a worldview;

- Multimodal receipt of information from the external environment;
- Multimodal delivery of information to the external environment;
- Control of movement organs and manipulators.

The listed set of functions is an extension of the set used in [53], however, such an extension is justified, since a deeper understanding of many aspects has emerged over the past time. If we really want to see artificial intelligence comparable to human intelligence, all these functions are necessary for it. As ten years ago, in a study by Alexey Samsonovich [54], all existing cognitive architectures were acclaimed as limited, we could not find more than 60% of the necessary functions anywhere. This served as a prerequisite for the development of a new cognitive architecture, which will have the necessary functionality.

In 2019, a cognitive architecture was proposed, which is a very general sketch of the AGI structure [55], annotated metagraphs were used to represent knowledge in the architecture [56]. Later in [57], an approach was proposed on how to create intelligent information systems based on this architecture, including individual components from the AGI composition. Due to the rapid progress of research in areas such as metagraphs, machine ethics, machine emotions, machine thinking, there is a need to reconsider this previously developed cognitive architecture. The need to develop a new cognitive architecture was the reason to reflect all the necessary functionality in the second version of this architecture.

## 2      Material and Methods

During the work, materials from several hundred scientific publications were used, the most significant of which are listed in the list of references. We often discussed which of several publications on the same topic to indicate in the link, and tried to bring the one that most fully reflects the information or more clearly sets out the material. Two publications in the same link were listed together under different numbers if they significantly complement each other.

Another important source of materials for the article was our own activities. After all, we are designing AGI, and it should be able to do a lot of things that a human can. Therefore, we analyzed our daily actions and listened to ourselves sensitively, trying to understand how we work with knowledge, and what functional components should be present as part of the cognitive architecture of AGI.

## 3      Results

In the process of the research, a preliminary design of a knowledge base has been developed that will be able to operate with any forms of knowledge representation. Using it, a preliminary design of a cognitive architecture has been developed, based on which AGI prototypes can be created.

### 3.1 The Universal Knowledge Model

#### 3.1.1 Archigraph as the Foundation of a Universal Model of Knowledge

The basis of cognitive architecture is the representation of knowledge. Like a human, an agent must be able to work with different forms of knowledge and switch from one to another. In [58], a universal data model was proposed that allows storing data in a data lake structured according to different data models: relational, multidimensional, graph, and others. For this purpose, the metagraph data model based on annotated metagraphs has been expanded through the use of protographs and archigraphs [59,60]. The annotated metagraph proposed in [61,62], but at first did not have such a name, is characterized by the following properties:

- The structure of the metagraph includes, in addition to the usual edges and vertices, metaedges and metavertices;
- Each vertex, edge, metavertex and metaedge is characterized by a set of attributes that have a name and value;
- Metavertices and metaedges differ from ordinary edges and vertices in that they can contain fragments of a metagraph inside themselves, which by their properties also represent metagraphs;
- The contents of various metavertices and metaedges can overlap up to complete equivalence (that is why the term annotated metagraph is used, that some meta-objects can annotate others [63]);
- The boundaries of the metavertices and metaedges are permeable to edges and metaedges to any nesting depth.

A protograph can be considered as a graph that has no edges. The role of edges is performed by the adjacent vertices to each other. The protograph P is defined by the set of elements $\{p_i\}$, $i = 1$, n and the neighborhood matrix $M_{n \times n}$ consisting of 0 and 1, where 1 means the neighborhood (adjacency) of element a to element b. Examples of protographs are: stack, queue, table, figures of the game "Life". An example of an infinite protograph is the tape of a Turing machine. Ordinary graphs are protographs in which elements are divided into two classes of elements and the rule applies that elements of the same class cannot be adjacent.

In the archigraphs, the elements belong to more than two classes. The number of classes into which the elements are divided is the most important characteristic of an archigraph. The concept of an archigraph makes it possible to systematize various definitions of metagraphs. Annotated metagraphs are archigraphs with five classes of elements: vertices, metavertices, edges, metaedges, and attributes.

To make it convenient to work with knowledge, we propose to expand the annotated metagraph model as shown in Fig. 1.

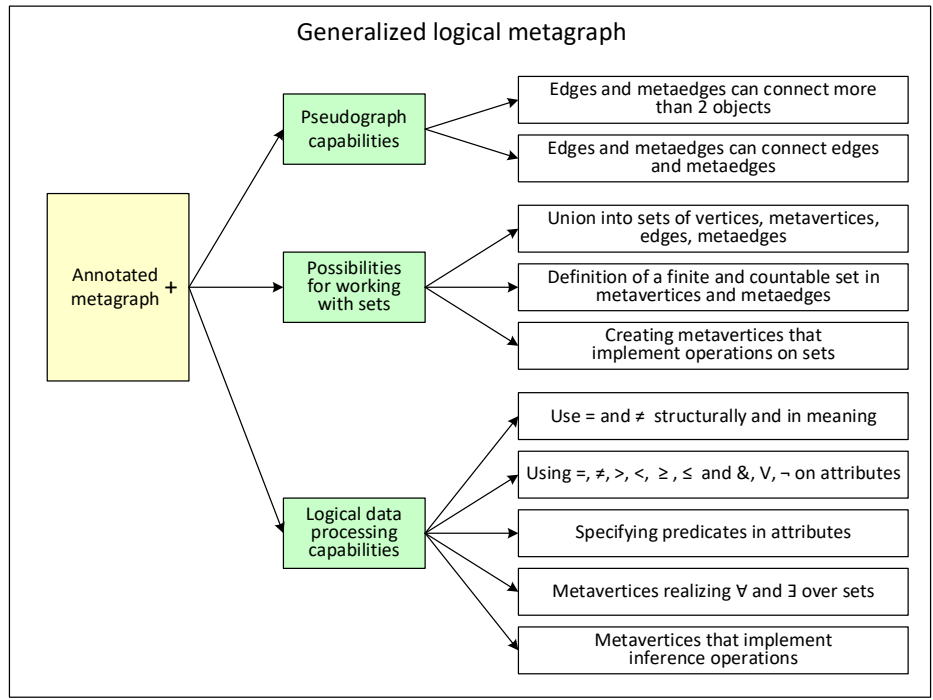

**Fig. 1.** An expansion of annotated metagraph.

This expansion of capabilities will be done through the following steps:

1. Add by analogy with pseudographs by A. Burdakov [64]:

— the ability to have edges and metaedges as in hypergraphs, which can connect more than 2 objects (can start on several objects and can end on several objects);
— the possibility that an edge or a metaedge can start not only from a vertices or a metavertices, but also from another edges or a metaedges, and can also end not only with a vertices or a metavertices, but also with another edges or a metaedges.

The term "pseudograph" used by A. Burdakov to describe the way knowledge is represented is not very successful, since there are at least two other interpretations of this term [65], [66].

2. Add features for working with implicitly and explicitly defined sets:

— to provide an opportunity to group vertices, metavertices , edges and metaedges into groups with similar ones due to their direct proximity to each other as in protographs. For any element of such groups, operations are possible: find out the number of elements in the group, get a link to the next member of the group;
— provide an opportunity for any metavertex or metaedge to indicate that the objects of the first level included in them form a finite set. For such sets, operations are possible: find out the number of elements of the set, get a link to the first nearest element of the set, get a link to the next element of the set;

— provide an opportunity for any metavertex or metaedge to indicate that the objects included in them form a countable set. For such metavertices or metaedges, two metavertices are created inside sets: one contains a finite set of object types, and the second contains a countable set of objects (further, when designing the knowledge base, various ways of representing countable sets will be proposed). For objects of a countable set, operations are possible: get a reference to the first nearest element of the set, get a reference to the next element of the set;

— provide an opportunity to create metavertices that implement operations on sets: union, intersection, subtraction, despite the fact that the internal elements of the first level of such metavertices can only be metavertices and metaedges that form sets.

3. Add logical data processing capabilities:

— to provide the possibility of operators = and ≠ above the objects in two versions: structurally (according to the internal structure and composition of attributes) and by value (in addition to the structure, the attribute values must match);

— provide the ability to use standard operations =, ≠, >, <, ≥, ≤ for operations on attributes;

— provide the ability to specify predicates in the attributes of vertices, edges, metavertices and metaedges;

— provide the ability to perform logical operations (&, ∨, ¬) over the attributes of vertices, edges, metavertices and metaedges;

— provide an opportunity to create metavertices that implement quantifiers ∀ and ∃ over sets, despite the fact that the only internal element of the first level of such meta-vertices can only be metavertices and metaedges forming sets;

— provide the ability to create metavertices that implement logical inference operations on objects, specifying a set of source objects and their attributes with predicates, as well as the object and its attribute to which the result will be assigned;

— create meta-vertices that implement a wide range of modal operators (aletic, epistemic, deontic, axiological, temporal) by specifying one or more objects and their attributes with predicates.

Metagraphs with such extensions can be used to implement first, second and higher order logic, as well as various modal logics, they only lack functions. Let's call them generalized logical metagraphs. To implement functions, we will move from metagraphs to archigraphs and in addition to vertices, edges, meta-vertices, and meta-edges, we will add another type of object - a function. The resulting generalized logical archigraphs can become the basis for the presentation of knowledge.

In order for the model to combine any forms of knowledge and ensure work with them, it is necessary, by analogy with the universal data model [58], to add several dozen types of elements corresponding to different forms of knowledge representation to the archigraph. Then it will be possible to build archigraphs in which all knowledge from a certain subject area will be combined, regardless of the form of their represen-tation. The edges and meta-edges of such an archigraph will correspond to the relationship between entities. Meta-edges will be used when these relationships are complex and need to be detailed with additional entities. In addition to edges and metaedges reflecting the content of knowledge, the archigraph will also contain edges and meta-edges reflecting the history of which object was obtained from or based on which. For example, that this text is an abstract of that text. The appearance of such seemingly technical connections in the archigraph is completely justified – this is also knowledge. The use of metaedges in such technical cases may be due to the complexity of the conversion process.

As a storage environment for the proposed archigraphs, it is necessary to develop a specialized DBMS that supports the archigraph data model. Its design and development can be considered as a further development of the work on the development of metagraph databases, approaches to the creation of which were considered in [67,68].

All the knowledge presented in the archigraph can be divided into three large groups: unformalized, partially formalized and formalized.

### 3.1.2 Storage and Processing of Non-Formalized Knowledge

The informal knowledge processed in computers can be classified as:

- Texts in natural languages, including technical, legal, prose, poetry, etc.;
- Various graph schemes and drawings developed outside the automation systems of their development;
- Maps made outside GIS;
- Speech audio recordings;
- Music, sound effects;
- Images (photographs, paintings, portraits);
- Videos and movies, including those with audio accompaniment.

There are many technologies for converting informal knowledge into formalized knowledge. For example, let's first consider the transition from natural language texts to formalized knowledge. In order not to delve too much into history, you can take as a starting point the popular and quite functionally complete NLTK library for Python [69]. During further development, neural networks and knowledge graphs began to be widely used for these purposes [70,71]. In addition to technologies for converting texts into semantic networks or into knowledge graphs (the latter provide opportunities for logical inference based on the knowledge contained in them), there are well-developed technologies for reverse conversion from graphs to texts [72]. One of the recent works in this field is devoted to the bidirectional transformation of texts into semantically loaded metagraphs and vice versa [73].

To formalize images, their representations are used in the form of scene graphs, which are structured representations of images in the form of graphs containing objects, their attributes and defining relationships between objects in the scene. There are currently two main approaches to scene graph generation (SGG):

- The first is to find objects, and then to find paired relationships between the found objects [74].
- The second is the simultaneous detection of objects and the relationships between them [75].

As a rule, SGG tasks are solved by using various types of neural networks: convolutional (CNN) [76], recurrent (RNN) [77], graph (GNN) [78]. [79,80] provides a detailed overview of existing methods for generating scene graphs. The reverse transformation – the generation of images from scene graphs, as well as from other images or from text descriptions is also performed using neural networks. Generative adversarial networks (GAN) are mainly used [81]. A separate major area is the generation of images and related descriptions based on the results of medical diagnostic procedures such as MRI and PET [82]. In these cases, the image acts as an external representation of the data processing results coming from the diagnostic equipment, masking the formalized representation of the semantics of the received data.

The task of formalizing video is relatively new. One of the solutions to this problem is based on the development of the idea of constructing a graph of image scenes – the construction of a graph of video scenes [83]. Another approach used in [84] involves the generation of knowledge graphs based on language annotations to videos. In the same article, the authors propose a model for generating knowledge graphs directly from video based on neural networks, trained on the dataset obtained in the first part of the work. Neural networks are also used for reverse conversion – generation of videos from knowledge graphs, sets of images or text descriptions. This topic attracts the attention of many researchers. Only in the preprint archive arxiv.org A search for the phrase "video generation" in the headlines of publications yields a list containing more than 380 articles. Some of the works on this topic are considered in the reviews [85], [86].

The formalization of audio recordings of speech is usually performed in two stages. Based on the spectral characteristics of speech, it is quite difficult to immediately build a knowledge graph. In order not to form it directly from the spectral pattern of human speech sound waves, an intermediate stage of speech recognition is performed, for example, using voice assistants [87]. After that, the problem is reduced to the task of formalizing texts in natural languages, which was considered earlier. Neural networks are widely used as speech-to-text translation tools: convolutional (CNN) [88], transformers [89] and, more recently, graph neural networks (GNN) [90]. About a dozen different methods have been developed for synthesizing speech from text [91]. Currently, neural network–based methods are also the most promising: convolutional networks that do not have autoregression provide the highest speed of speech formation [92], and networks with feedback - transformers [92] and generative-adversarial (GAN) [93] are characterized by high acoustic speech quality, can generate speech with several voices and give it an emotional coloring.

To formalize musical works, their fragments and the sounds of musical instruments stored in musical databases [94,95], as well as sound effects stored in special databases [96], special formalization methods based on Markov processes [97] and algebraic methods [98,99] are used.

A separate major area of formalization of non-formalized knowledge is the formalization of multimodal representations combining two or more non-formalized streams of knowledge, such as those related to the main ones: text, image, video, audio, as well as various auxiliary ones: context, pose, intonation, facial expressions, smell, taste, touch. The knowledge graph obtained as a result of multimodal synchronous processing of several parallel streams of unformalized knowledge is not a simple combination and combination of knowledge graphs obtained by processing individual streams, additional knowledge may appear in it due to a deeper understanding of the subject area [100]. In addition, knowledge from different streams fills in gaps and corrects errors in individual streams. It is shown that the connection of several modalities makes it possible to improve the formation and processing of knowledge graphs created on the basis of information selected from social networks [101]. Interactive immersive generative multimodal interaction between a person and an agent in the form of a steady smooth exciting conversation accompanied by the display of images is considered as a prospect for the development of technologies of multimodal interaction with agents in [102].

### 3.1.3   Storage and Processing of Partially Formalized Knowledge

The first group of partially formalized knowledge is data that has a structure, but there is no intensional that allows them to be used and interpreted. And there is also no understanding of the place of specific data in the metric of the relevant semantic space. Partially formalized knowledge can include:

- Data located in files organized using various access methods (sequential, direct, index-sequential, etc.);
- Data stored in files created without specifying access methods in various local, distributed and cloud file systems;
- Data in databases organized according to various data models (network, hierarchical, multivalue, multidimensional, relational, object, vector, XML, key-value, wide column, documentary, tabular, time series, event, spatial, etc., as well as RDF used without describing semantics) [103], including: data from various modules and subsystems of enterprise management, data from document management and content management systems, data from electronic trading platforms, test questions and training material from automated learning systems, data from library systems and scientific citation systems, data from research automation systems, etc.;
- Data from search engines indexes, both on the internet scale [104] and corporate ones [105];
- Data from blockchain frameworks;
- CAD data on the products being designed;
- Neural network structures;
- Cartographic data in GIS;
- Tables and diagrams prepared in desktop and cloud applications, with the exception of diagrams describing certain sequences of actions (program flowcharts, business process diagrams, project plans, production flowcharts, etc.).

The second group of partially formalized knowledge is mathematical models that are not context–bound:

- Linear and nonlinear equations and systems of such equations;
- Differential equations and systems of such equations;
- Partial differential equations and systems of such equations;
- Probabilistic equations and systems of such equations;
- Logical equations of propositional calculus and systems of such equations;
- Tensor operators and equations;
- Infinite-dimensional topological vector spaces and their mappings [106];
- Algebras, groups, rings, fields, lattices, modules [107].

For both groups, the lists are clearly not exhaustive. However, they provide an understanding that allows you to assign similar cases, that are not included, in each group.

To turn data or abstract mathematical models into formalized knowledge, it is necessary to add connections that allow them to be used and interpreted, to correlate these objects with other ones.

### 3.1.4   Storage and Processing of Formalized Knowledge

The ways of presenting formalized knowledge are very diverse:

- Computer programs in traditional programming languages (programs that have source code, architecture, or algorithm descriptions available will have internal granularity);
- Computer programs in languages that implement logical programming based on a subset of first-order predicate logic or implement it along with other features (Prolog, Visual Prolog, Mercury, Oz, Strand, KL0, KL1, Datalog, etc.) [108];
- Diagrams describing some sequences of actions (program flowcharts, business process diagrams, project execution plans, production flowcharts, etc.) Prepared using specialized tools;
- Semantic networks [109];
- Trained neural networks;
- Frames (during the first 15 years of their development, frame systems and languages were used for the structural representation of knowledge [110], but since the late 1980s, logical inference tools based on stored information have appeared in them [111] and with subsequent development, the role of logic tools increases significantly [112]);
- Knowledge graphs [113];
- Descriptions using first-order predicate logic (in the languages of general logic [114], as well as in the languages CycL [115], FO[·] [116], KIF [117,118], etc.);
- Descriptions using language families for the semantic web: RDF [119] and OWL [120] or simpler ones like SHOE [121];
- Description in ontology description languages and/or in systems such as UFO [122], OntoUML [123], DOGMA [124], Ontologua [125], LOOM [126], etc.;

- Production systems (in simple cases, they can be implemented using traditional programming languages, with professional implementation specialized development languages [127,128] or special systems for working with production rules [129] are used, they can be divided by type into clock and stream ones [130]);
- Formal grammars [131];
- Formal systems of concepts [132];
- Conceptual models of knowledge [133];
- Interconnected points in the Elements-Attributes-Relations space [134,135];
- Complex networks [136];
- Petri nets [137];
- Finite state machines [138];
- Simulation models [139].

Traditionally, when listing ways to represent formalized knowledge, objects such as computer programs, neural networks, finite state machines, simulation models and complex networks are not considered. This is due, among other things, to the fact that in the absence of additional information, and sometimes even if it is available, it is impossible to explain the results obtained by accessing such objects. However, when creating a broad-purpose knowledge base for a projected AGI, the lack of detailed explanations of how the final result was obtained, as well as for a person, is not a reason to discard such objects. We can say that they contain "canned" knowledge, and when entering the initial data, they produce a result.

The above list of types of formalized knowledge is also not comprehensive, as is the composition of other groups, but it gives an intuitive understanding of what objects, not yet listed, could be included in this group.

### 3.1.5 An Alternative Version of the Knowledge Base Organization, which was Abandoned

It was possible to avoid applying the concept of an archigraph with several dozen types of objects for different types of knowledge, but instead to focus on the model of a generalized logical metagraph, to which only functions could be added and a "Type" attribute could be introduced for each vertex, with which the type of knowledge contained in it could be determined. But for each type of knowledge, its own group of functions working with it will be used, which may overlap or not overlap with the functions of the other types of knowledge. In addition, technical edges or metaedges related to knowledge transformation would come into or out of each object. And for all these functions, edges and metaedges in this case, one would also need a special "Type" attribute. This would complicate the structure. Therefore, this option was abandoned.

## 3.2    A Preliminary Design of the Cognitive Architecture on the Basis of which AGI Prototypes Can Be Developed

### 3.2.1    Common Description of Cognitive Architecture

Based on the suggested universal knowledge model, a cognitive architecture is proposed, shown in Fig. 2, which can be used to develop AGI prototypes.

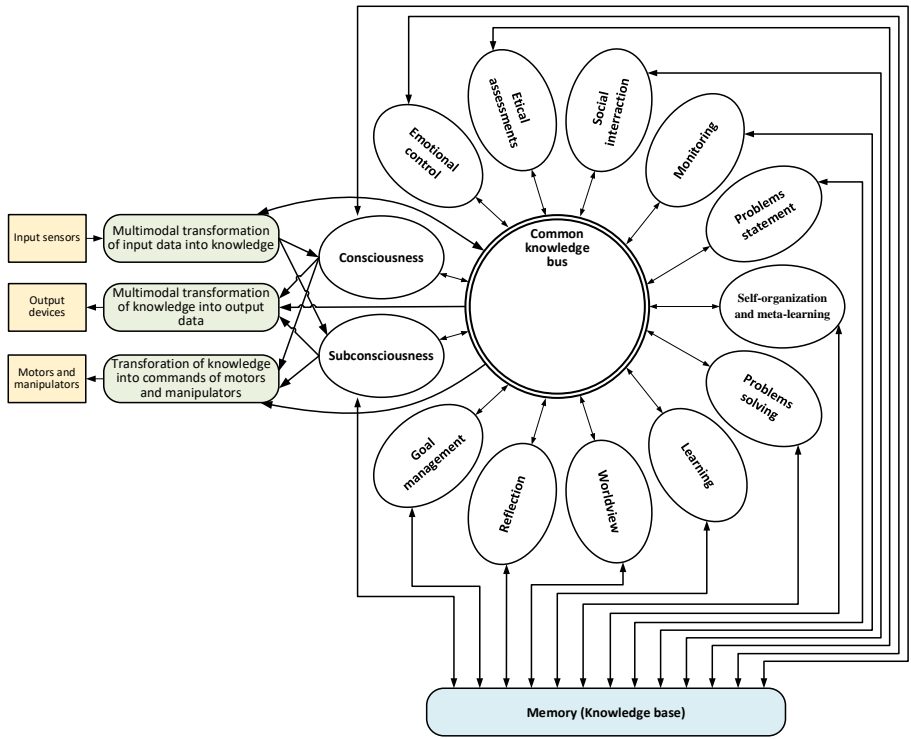

**Fig. 2.** A diagram of the cognitive architecture that can be used to develop AGI prototypes.

All functional modules included in the cognitive architecture, except for modules interacting with the external environment, work with the knowledge base, which is the agent's memory. Each module interacting with the knowledge base has a separate section there, in which it stores the knowledge it needs, and can change them as it sees fit. If some knowledge is important for several modules, to eliminate duplication, they can be stored in a common section accessible to all of them. But then, to make changes to such knowledge, consent must be obtained from all modules that have access to them.

All functional modules of the cognitive architecture are integrated using a common knowledge bus. The knowledge sent to the common bus gets to all modules connected to the bus, then each module individually decides how to deal with them, save them in its knowledge base area and somehow use or ignore them. In addition, the shared bus can be used by modules to request knowledge from other modules. The response to the request also comes via a common bus. Knowledge on the common bus can be transmitted in all forms in which it can be stored in the knowledge base.

Using a shared knowledge bus is an alternative used in many cognitive architectures to the concept of working memory ([9], [11], [12], [15] and others). These options are comparable to the integration methods used in the architecture of software systems: integration over a common bus and integration over data. The advantages of using a common knowledge bus are that all modules are forced to process the incoming knowledge stream in parallel, except in situations where knowledge is sent to a specific module in response to its request. This is a mechanism that requires much more resources (both processor and memory), but this mechanism is based on active actions, it involves individual modules in knowledge processing and provides an up-to-date context for all of them. Unlike the general knowledge bus, when using working memory, in case of any situation, the module must look at what lies in the workspace. This is a passive mechanism. What if the necessary knowledge has already been removed from the workspace and replaced with something else?

The modules of consciousness and subconsciousness play a major role in interacting with interface modules that provide interaction with the external environment. The consciousness module is a control system that controls the movement of an agent in a complex space, which is a combination of ordinary three-dimensional space, time and any other spaces in which the agent operates (for example, media spaces, consumer goods spaces, scientific research spaces, etc.), with all the limitations inherent in these spaces (for example, you can only move from one floor to another by stairs or by elevator). The subconscious module allows you to implement ready-made routine sequences of actions stored in the knowledge base without connecting or with minimal connection of a relatively slow control system implemented by consciousness. Often the same sequences of actions performed by consciousness move over time into the subconscious.

Consciousness controls movement, focusing on the goals generated by the goal management module, and on the situation developing in the external environment. Knowledge about the situation in the external environment from the multimodal data and knowledge conversion module is transmitted not only to the modules of consciousness and subconsciousness, which are the main consumers this data, but also after being cleared of unnecessary details, it is transmitted to the common bus, from where all other modules receive them, including: monitoring module, goal management module, emotional management module. And the same in the opposite direction (but not symmetrically): the modules interacting with the external environment receive knowledge from the common bus, for example, emotions from the emotional management module, although the bulk of knowledge is transferred there from consciousness and subconsciousness.

The emotional control module is another control system besides consciousness, which does not act as precisely as consciousness can, but on all modules at once. The emotions generated are not some single objects with the meanings "sad", "joyful", "anxious", "ashamed", etc., but more complex knowledge structures where there are reasons, some previous history, perhaps: participants in the event and something else.

The ethical assessment module generates ethical assessments of all events and actions, including possible actions. These assessments can both encourage and deter these actions, and the level of support or deterrence may vary. If the actions and events have already taken place, then their ethical assessments are added to their characteristics. Just like emotions, ethical assessments are complex structures in which reasons may be present, justifying artifacts or chains of events, etc. Ethical assessments are formed on the basis of ethical principles stored in the knowledge base and amenable to modification either with human participation or as a result of repeated occurrence of sequences of events that reveal contradictions in these principles.

The reflection module is somewhat similar to the ethics module, it also generates assessments of actions and events, but only past ones, and not from the perspective of some ethical norms established by culture, religion or some authorities, but from the perspective of achieving the final result, side effects that arise, influencing one's own and others' plans, and etc. At the same time, very often, during the work of the reflection module, events that have occurred are modeled from the positions of other participants, their assumed (or confirmed by some facts) estimates of these events are formed, and their own estimates are adjusted from these positions.

The social interaction module takes into account relationships with people and other agents, existing and possible roles in these interactions, emotions towards other agents, the course of processes and the results of previous interactions with them and, based on all of this, adjusts action plans that are somehow related to other agents. In addition to plans, the module influences actions and dialogues when interacting with people and other agents, forming additional knowledge for modules of interaction with the external environment.

The worldview module provides the formation and support of a special section in the knowledge base, which contains a picture of the agent's world and determines his place in it, the main goals of the agent's existence are formulated. This section is based on a scientific picture of the world, but to understand many aspects of history, art, and social relationships, other pictures of the world (mythological, several religious, and alternative scientific) are also stored in the section with explanations why they are not correct. The scientific picture of the world is modified as new scientific knowledge becomes available. Stored worldviews are used to gain new knowledge from different fields and compare them with those available in the database. In case of contradictions, various reactions are possible, both the re-clarification of the received knowledge and the assignment of assessments that they are false.

The monitoring module allows for various types of monitoring in a wide range of processes occurring in the external environment and in the modules of the agent's cognitive system. The occurrence of some events, changes in the values of the characteristics of some objects can be controlled, and both the achievement of some expected value and the finding of this value within some acceptable limits can be controlled. The initiation of monitoring processes is carried out by other modules of the cognitive architecture.

The learning module generates new knowledge in the knowledge base. This can happen both through the direct transfer of knowledge coming from the external environment with their subsequent comparison with existing knowledge and binding in the absence of contradictions, and through the formation of models based on the results of the agent's own actions. Both externally received and independently created models are tested and checked for compliance with the knowledge already available in the knowledge base. If there are inconsistencies, new knowledge can be discarded (if it strongly contradicts the worldview, ethical assessments, or already existing knowledge) or recorded in the knowledge base with the mark indicating the presence of contradictions. The formation of new knowledge can occur in all modules of the cognitive architecture working with the knowledge base, while access to the learning module is performed to replenish and/or adjust the knowledge base.

The task setting and solution modules carry out the formulation and solution of tasks. Requests to them are initiated from any module of the cognitive architecture, including the learning module for building new models and the setting and solving problems modules themselves for setting and solving problems when a complex task is divided into subtasks.

With the help of the self-organization and meta-learning module, the agent should be able to rebuild and improve his activities, find and switch to using new more productive ways of learning. Decisions on the reorganization of activities or on the transition to new forms of activity, as well as decisions on the transition to new forms of education should be made based on the results of a purposeful search, modeling and subsequent practical testing. For the first time, the need for such a functional component as part of the cognitive architecture for AGI was shown in [140].

The presented description of the cognitive architecture in the light of [141] can be considered as a top-level specification that can be used to further detail the cognitive architecture and develop an intelligent agent, however, we will not delve into this process due to its resource intensity. Our following descriptions of the individual modules will be related to an overview of the implementation of the corresponding mechanisms in other cognitive architectures.

### 3.2.2 Consciousness

Of the 42 cognitive architectures considered by us, only four have an explicitly highlighted "Consciousness" component for building AGI: LIDA [12], Haikonen cognitive architecture [13], [16], ICOM [23] and MBCA [27], Clarion [47], OntoAgent [49], ISAAC [51], [52]. And GLAIR [31] also talks about the Knowledge Layer in which conscious reasoning, planning, and act selection is performed. There are much more cognitive architectures for simpler robots that have consciousness embedded in them, or architectures being developed as part of research projects on the realization of consciousness: [142], [143], [144], MECA [145], [146], [147], [148], [149], RoboErgoSum [150], [151], COCOCA [152], MLECOG [153], [154], [155], CELTS [156], [157], [158], [159], [160], [161]. In these works, as a rule, a lot of attention is paid to the technical implementation of consciousness and its integration into the cognitive architecture, while its functionality is not worked out deeply enough, but only at a general level.

The key features of the modern understanding of machine consciousness were formulated in [162,163]. Consciousness is considered as a control system for the agent's current actions. To have minimal consciousness, it is necessary that the agent has the following capabilities:

- self-knowledge: Agent has complete knowledge of its current cognitive state as well as of the data produced by all its interfaces, sensor, and motor units.
- self-monitoring: Agent is completely informed about the performance and status of its sensory and motor units over time (including the quality of the sensations and the reports from all of them) and of its embedding in the environment as it is.
- self-awareness (or self-reflection): Agent behaves in a way that unambiguously reflects, respectively is determined by its current cognitive state and the information gained by its self-knowledge and self-monitoring abilities, and that is 'aware' of the internal and external changes that it causes.
- self-informing: Agent globally broadcasts its cognitive state, to all modules of the system and whenever changes of state occur.

It can be expected that the further development of this theory from a minimal to a more advanced consciousness will make it possible to realize consciousness as a system of parallel control processes taking place in multidimensional, partially intersecting virtual spaces in which spatial and temporal constraints are set. If the actions that need to be performed based on the results of parallel management processes in the real space in which the agent operates contradict each other, they are checked for consistency in time distribution, ranked and queued. In the process of functioning, the consciousness module intensively interacts with all other AGI modules, receiving plans for further actions, applying for statements and solutions to problems of modeling possible situations, considering ethical assessments, checking with the worldview, etc.

### 3.2.3 Subconscious

The machine subconscious contains ready-made models and algorithms that can be quickly activated when corresponding situations arise. Unlike human memory, the volume of the machine subconscious can be quite large, and stored models can cover a wide range of fields of knowledge and activities. Among the architectures designed to create AGI, the following architectures have the subconscious: ACT-R [10], LIDA [12], CogPrime [17, 18], ICOM [23], GLAIR [31], MicroPSI [40], Clarion [47], OntoAgent [49], ISAAC [51], [52]. Some cognitive architectures, which do not have the immediate goal of creating AGI, are claimed to have both consciousness and subconsciousness: COCOCA [152], MLECOG [153], [154], [155], CELTS [156], [157], [158], [159], [160], [161]. At the same time, there are cognitive architectures of this class, which have only subconsciousness without consciousness: IFORs [164], [165], [166], [167], DIARC [168], NCCA [169], [170].

The existence of the subconscious in the cognitive architecture is evidenced not only by the explicit indication of its presence. As was rightly noted in [35], all reactive cognitive architectures such as Soar [9], ICARUS [171], SW-CASPAR [172] or individual reactive levels in complex combined architectures, such as the reactive layer in CogAff [34], contain a certain set of behaviors for different situations and, in fact, act as a subconscious.

The perception of subconsciousness, that has developed among some researchers, as some kind of auxiliary system that helps consciousness (which is not entirely true) leads to the fact that the specific functionality of the subconscious in cognitive architectures is often left behind the scenes or distorted. So, in [155] and [157] they take as a basis Kahneman's idea [173] about the dual nature of thinking processes and distinguish a fast subconscious mind that processes large amounts of incoming information and a slow consciousness that processes significantly less information. This is true, but with this approach, the functionality of the subconscious mind for the accumulation, storage and use of ready-made algorithms and models remains undisclosed.

Along with the fact that the functionality of the subconscious mind is not revealed, it can also be distorted. So, the article [174] begins with the correct statements that the subconscious can handle tasks in the high dimensional problem solving space while the consciousness can operate only in the low dimensional space. But then the author considers a way to transfer information from the subconscious mind to consciousness using emotions. The same idea is expressed in [175]. In our opinion, this is wrong, emotions are not signals of the subconscious mind. The emotional management system is a separate functional component of the cognitive architecture. In the architecture shown in the figure above, the subconscious mind can directly interact with the modules of interaction with the external environment. Consciousness, receiving information about these interactions, can suppress or correct them.

Similarly, attaching the functions of interaction with the external environment to the subconscious mind [161], in our opinion, is also a distortion of its functionality; for this purpose, separate modules should be allocated in the cognitive architecture. The restriction on the methods of realization of the subconscious mind caused by this distortion (the use of neural networks and fuzzy logic) immediately becomes invalid after the distortion is eliminated. This is confirmed by a number of examples of the inclusion of the subconscious mind in the symbolist cognitive architectures presented in this section, based on the use of production rules.

### 3.2.4 Worldview

Of the 42 cognitive architectures we have considered for building AGI, the worldview module is present in NARS [11], Sigma [15], MBCA [27], GLAIR [31], H-CogAff [34], EM-ONE [36], CAMAL [38], SMCA [39], Clarion [47] and Oscar [48], as well as in cognitive architectures developed for research purposes that did not consider all aspects of human activity: CogAff [34], [145], ICARUS [171], SW-CASPAR [172], CRIBB [176], A-CRIBB [177], [178], SACA [179], [180], CoJACK [181], [182], [183], MAMID [184], CogToM [185], Scruff [186], CASPAR [187], [188] and InnovA [189]. All these cases provide for the presence and use of many elements of worldviews. All these elements are either intuitive (laid down initially when creating an agent), or formed as a result of training, and they are all related to some operational aspects of the activity. Conflicting worldviews such as scientific, religious, and mythological are not considered.

In [190], a method for the coordinated use of several contradictory worldviews is proposed, but it is quite primitive and not suitable for AGI: the agent acts as a character in a game journey full of challenges and mysteries that underlie myths. In [191] and [192], a fragmented worldview is considered. The fundamental basis for dealing with worldviews is the theory of Belief revision [193], [194], [195]. For effective coordinated use of different worldviews, it is necessary, depending on the context, to ensure dynamic switching between different independent worldview models and to have knowledge (on a deeper layer of worldview common to all models) about the relevance of each model and the conditions of its application.

### 3.2.5 Reflection

The reflection module builds models of this agent, reflecting its various aspects: activities, plans, knowledge, appearance, etc. At the same time, these models can be built in parallel and from different perspectives:

- From the point of view of some theoretical concepts: cost, safety, environmental protection, technical condition, etc.;
- From the point of view of some real or abstract individuals and/or organizations. In the process of building a reflexive model, an agent model is used, from which position this reflexive model is built. In this case, not one, but a small sequence of reflexive models can be built from the position of a particular agent and several of its models: In particular, the second reflexive model is built on the assumption that the external agent knows that the main agent understands what the external agent thinks about him. This is considered in the second external agent model. According to the same principle of multiple reflection, third and subsequent reflexive models and models of an external agent can be constructed.

In both variants, emotional assessments of concepts and subjects can be considered, from the point of view of which reflexive models are built.

The created reflexive models can be partial models of a given agent (reflect only one or several of its sides) and differ from the model of a given agent available in consciousness. They are used for self-assessment and adjustment of modelled aspects.

Reflection functionality is available in the following cognitive architectures designed to create AGI: Soar [9], Sigma [15], eBICA [21.22], H-CogAff [34], EM-ONE [36], CAMAL [38], SMCA [39], MicroPSI [40], [44], Polyscheme [45], [46], Clarion [47]. In addition, it is used in the cognitive architecture of strong human-machine intelligence [196], as well as in cognitive architectures designed for simpler robots and research in the field of cognitive architectures: CogAff [34], DIARC [168], CRIBB [176], A-CRIBB [177], SACA [179], [180], MAMID [184], [197], EM1 [198], Cognitive Architecture for Human-Robot Interaction [199], [200], [201], [202].

## 4    Conclusions

The article proposes a model of knowledge representation that allows the upcoming AGI to work with different representations of knowledge and freely move from one to another. This model is used in the proposed AGI cognitive architecture, which identifies functional blocks for all types of cognitive activity that the authors considered inherent in humans.

We hope that the proposed cognitive architecture and knowledge representation model will solve the problem formulated in the introduction, as well as allow us to answer some of the challenges that faced the theory of cognitive architectures according to [203,204], in particular:

- The need for robust integration of mechanisms involving planning, acting, monitoring and goal reasoning.
- The limited size and the homogeneous typology of knowledge that is encoded and processed by systems based on cognitive architectures.

In the first prototypes being developed, not all types of knowledge and not all functional blocks of the cognitive architecture may be supported. At the same time, the prototyping process and future theoretical research may reveal the need to expand the composition of blocks of cognitive architecture, to include blocks not provided for in this article.

# 5    Discussion

The article considers in detail only 5 functional modules of the proposed cognitive architecture out of 17 including Knowledge base. Of course, it is necessary to describe them all. Obstacles to this:

- the text of the article is already quite large then will increase significantly in size;
- a lot of additional time is required to perform a significant additional amount of work on the analysis of cognitive architectures and text preparation.

The authors plan to prepare a continuation of this article in the future with a description of the remaining functional modules.

# 6    Organizational Snippets

## 6.1    Author Contributions

**Artem Sukhobokov**: Conceptualization, Writing - Original Draft, Supervision. **Evgeny Belousov**: Resources, Writing - Original Draft. **Danila Gromozdov**: Writing - Original Draft, Validation. **Anna Zenger**: Writing - Original Draft, Writing - Review & Editing. **Ilya Popov**: Methodology, Writing - Original Draft.

## 6.2    Declaration of Competing Interest

The authors declare that they have no known competing financial interests or personal relationships that could have appeared to influence the work reported in this paper.

### 6.3   Funding

This research did not receive any specific grant from funding agencies in the public, commercial, or not-for-profit sectors.